\documentclass[a4paper]{article}
\usepackage[utf8]{inputenc}
\usepackage[T1]{fontenc}
\usepackage{amsmath,amssymb,array}
\usepackage{algorithm}
\usepackage[noend]{algpseudocode}
\usepackage{booktabs}
\usepackage{xcolor}
\usepackage{graphicx} 
\usepackage{wrapfig}
\usepackage{float}
\usepackage{amsthm}
\theoremstyle{definition}

\usepackage{amssymb}
\usepackage{hyperref}
\usepackage{xspace}
\hypersetup{
    colorlinks,
    linkcolor={red!50!black},
    citecolor={blue!50!black},
    urlcolor={blue!80!black}
}
\usepackage{todonotes}
\usepackage{natbib}
\usepackage{arxiv}
\graphicspath{ {./images/} }

\definecolor{cverbbg}{gray}{0.93}

\newcommand{\Xmetatest}{  X_{meta-test}}
\newcommand{\XmetatestSp}{  \mathcal{X}_{meta-test}}

\newcommand{\YmetatestSp}{  \mathcal{Y}_{meta-test}}

\newcommand{\DmetatestSp}{  \mathcal{D}_{meta-test}}

\newcommand{\XmetatrainSp}[1]{  \mathcal{X}^{#1}_{meta-train}}
\newcommand{\XSp}{  \mathcal{X}}

\newcommand{\YmetatrainSp}[1]{  \mathcal{Y}^{#1}_{meta-train}}
\newcommand{\YSp}{  \mathcal{Y}}
\newcommand{\DSp}{  \mathcal{D}}

\newcommand{\DmetatrainSp}[1]{  \mathcal{D}^{#1}_{meta-train}}
\newcommand{\Sone}{S1\xspace}
\newcommand{\Stwo}{S2\xspace}
\newcommand{\Sthree}{S3\xspace}
\newcommand{\Sfour}{S4\xspace}

\DeclareMathOperator*{\argmax}{arg\,max}

\usepackage{caption}
\begin{document}


\title{Consolidated learning - a~domain-specific model-free optimization strategy with examples for XGBoost and~MIMIC-IV}

\author{
    Katarzyna Woźnica \\
    Faculty of~Mathematics and~Information Science \\
    Warsaw University of~Technology\\
    \texttt{k.woznica@mini.pw.edu.pl} \\  
    \And
    Mateusz Grzyb \\
    Faculty of~Mathematics and~Information Science \\
    Warsaw University of~Technology\\
    \texttt{m.grzyb@student.mini.pw.edu.pl} \\
    \And
    Zuzanna Trafas \\
    Faculty of~Computing and~Telecommunications \\
    Poznan University of~Technology \\
    \texttt{zuzanna.trafas@student.put.poznan.pl} \\
     \And
    Przemysław Biecek \\
    Faculty of~Mathematics and~Information Science \\
    Warsaw University of~Technology \\
    \texttt{przemyslaw.biecek@pw.edu.pl} \\
}

\maketitle


\begin{abstract}
   For many machine learning models, a~choice of~hyperparameters is a~crucial step towards achieving high performance. Prevalent meta-learning approaches focus on obtaining good hyperparameters configurations with a~limited computational budget for a~completely new task based on~the~results obtained from the~prior tasks. This~paper proposes a~new formulation of~the~tuning problem, called \textit{consolidated learning}, more suited to~practical challenges faced by model developers, in which~a large number of~predictive models are created on similar data sets. In~such settings, we are interested in the~total optimization time rather than tuning for a~single task. We show that a~carefully selected static portfolio of~hyperparameters yields good results for anytime optimization, maintaining ease of~use and~implementation.
Moreover, we point out how to~construct such a~portfolio for specific domains. The~improvement in the~optimization is possible due to~more efficient transfer of~hyperparameter configurations between similar tasks. We demonstrate the~effectiveness of~this approach through an~empirical study for XGBoost algorithm and~the~collection of~predictive tasks extracted from the~MIMIC-IV medical database; however, \textit{consolidated learning} is applicable in~many others fields. 
\end{abstract}


\maketitle

\section{Introduction}\label{sec1}
 In order to~effectively use the~full capabilities of~available machine learning algorithms, we have to~pay considerable attention to~the~hyperparameter settings. On the~one hand, hyperparameter tuning may be very expensive due to~the~dimension of~the~searched space. On the~other, it is necessary because the~default settings of~the~hyperparameters do not guarantee a~good model quality~\citep{lavesson2006quantifying,probst2019tunability}. So, automatic hyperparameter optimization methods are being developed to~avoid a~manual, trial-and-error-based search for the~optimal set and~thereby support users in building effective predictive models. They have become part of~AutoML frameworks ~\citep{autoweka2013,hyperopt2015,tpot2016,feurer2019auto} and~resulted in increasing attention to~the~ease of~use, implementation, parallelization, and~computational complexity of~the~proposed methods. It is essential to~adapt to~the~considered prediction problem and~provide anytime performance, i.e., to~propose a~good configuration of~hyperparameters even if only a~few evaluations have been performed.

    So far, two main groups of~optimization techniques have been recommended and~are used as~baselines in papers proposing new solutions. The~most basic class of~methods are grid search and~random search ~\citep{randomSearch2012}. They are completely independent of~the~dataset; for each case,  optimization must be started from scratch for a~pre-specified hyperparameter grid. To~find near optimal solutions, many optimization evaluations have to~be performed. In addition, these methods do not use the~information obtained in the~earlier iterations, namely the~information which algorithm settings gave a~good performance model. The~second class, Bayesian-based methods, is a~response to~that problem. It automatically extracts knowledge from the~learning curve, and~then the~surrogate model proposes a~new set of~hyperparameters weighing the~benefits of~exploring new, unseen regions versus sampling from the~known regions with good performance~\citep{hutter2011sequential,bergstra2011algorithms,snoek2012practical}. This is an~example of~online hyperparameter optimization adapting to~dataset characteristics and~updating the~learning curve. Nevertheless, these methods still do not provide anytime performance and~require independent optimization for each prediction problem.
    
In addition to~the~techniques that require performing a~full optimization for each new task, there is an~increasing need for an~offline approach that involves building a~portfolio of~several hyperparameter configurations ~\citep{Wistuba2016SequentialTuning,feurer2019auto,pfisterer2021learning,feurer2021autosklearn}.   From one perspective, the~hyperparameters portfolio is a~special case of~meta-learning since the~portfolio components together should  give good performance on previously performed experiments for the~collected datasets and~should transfer this good performance to~a~new dataset. For each collected dataset, at least one configuration from the~portfolio should parameterize the~model with a~good quality. We assume that at least one configuration will be promising for new, unknown data. Such a~repository of~data based on what we determine as~the~portfolio is called a~meta-train set and~new target predictive problems are called a~meta-test.

It has been shown that a~predefined, limited set of~hyperparameters optimized for a~wide range of~datasets gives better results than Bayesian optimization~\citep{Wistuba2016SequentialTuning,pfisterer2021learning}. Moreover, the~portfolio approach may be seen as~extended defaults that are easy to~share and~parallelize. In the~first studies introducing this method all meta-train datasets have the~same relevance while composing a~portfolio due to~their independent weighing. Therefore, to~enhance the~impact of~meta-learning, \cite{feurer2019auto} use meta-features, (i.e. vectors of~dataset characteristics) in the~evaluation of~the~dataset similarity. Then, while creating a~portfolio, more weight is given to~good hyperparameters configuration for meta-train sets more similar to~the~considered new data. This approach may be seen as~a~combination of~the~online and~offline procedures since a~static portfolio leverages the~most effective configuration for similar datasets, assuming they have similar learning curves. Employing meta-train datasets allows reduction of~time for early iterations in Bayesian methods.

Techniques employing portfolios built on meta-features and~assessing similarity are intuitive to~humans and~resemble an~expert's use of~domain knowledge. The~difficulty of~this approach is actually the~use of~meta-features. Firstly, computing meta-features may be expensive and~generate errors \citep{feurer2021autosklearn}. Secondly, we do not know how to~describe prediction problems and~datasets using meta-features in an~effective and~discriminative way. Namely, whether they should be predefined, based on statistical definitions~\citep{Vanschoren2019,Rivolli2018TowardsMeta-learning}, or landmarkers ~\citep{pfahringer2000meta}, or perhaps automatically trained extractors based on neural networks~\citep{edwards2016towards,hewitt2018variational,jomaa2021dataset2vec}. Due to~availability, a~set of~meta-features based on statistical definitions is most often used, but their correlation with model performance is questionable~\citep{woznica2021explainable}. Also, the~definition of~distance and~similarity between data sets is underdetermined~\citep{wistuba2015learning,feurer2015initializing}.

In addition to~meta-features,  selection of~meta-train is crucial for the~effectiveness of~the~portfolio. The~standard choice of~dataset repositories is OpenML~\citep{bischl2017openml}. It includes prediction problems from diverse domains and~may be a~satisfying source to~build a~portfolio that speeds up the~optimization for general, random data.  Nonetheless, we have some external knowledge about specific characteristics in many applications, e.g., a~high target imbalance in insurance claims frequency models or interactions between specific blood tests in medical data. Domain-specific AutoML frameworks~\citep{alaa2018autoprognosis,chalearn2019analysis,vakhrushev2021lightautoml} already employ these unique properties. This work shows that instead of~searching for meta-features describing these relevant attributes we can appropriately select the~meta-train, limiting to~representative datasets from a~specific domain. We call this refined meta-train a~\textit{consolidated meta-train} and~we term the~subsequent creation of~a~portfolio to~transfer hyperparameter from that meta-train as~\textit{consolidated learning}.

    \textbf{Our contributions } are as~follows. 1)~We purposefully restrict meta-train distribution, taking into account domain-specific characterizations of~considered tasks. Defining a~\textit{consolidated meta-train}, we highlight the~importance of~design decision in selection of~meta-train.   

\begin{figure}[h!t]
    \centering 
    \includegraphics[width=0.9\textwidth]{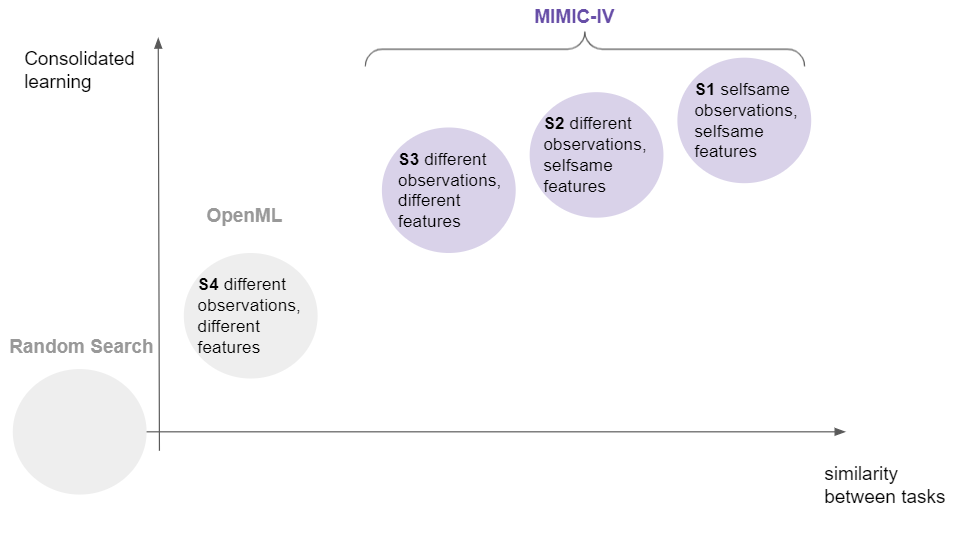}
    \caption{Relationship between similarity of~tasks and~\textit{consolidated learning}. Correspondence between space of~prediction problems allows meta-feature-free technique of~hyperparameter tuning to~incorporate the~advantages of~meta-learning. The~more similarity in the~design of~the~tasks, the~ \textit{consolidated learning} increases.}
    \label{fig:diagram}
\end{figure}

    2)~We leverage a~consolidated collection of~the~prior experiments to~determine the~portfolio of~hyperparameters transferred from the~meta-train to~meta-test tasks. We employ two model-free portfolio selection strategy methods: greedy search and~average ranking. 
    3)~To mimic a~real case, we create a~ metaMIMIC repository extracted from the~medical MIMIC-IV database~\citep{johnson2020mimic}. Our~experiments reflect various levels of~consolidation between the~meta-train and~the~meta-test (see~Figure~\ref{fig:diagram}) because of~the~definition of~input and~output space for every task.
    4)~In our experimental setup, we empirically show an~improvement of~\textit{consolidated learning} on baseline methods (random search and~Bayesian optimization) and~predefined portfolios extracted from the~OpenML repository.   We confirm the~hypothesis that \textit{consolidated learning} for MIMIC-IV enhances the~transfer of~XGBoost~\citep{xgboost} hyperparameters in the~early stage of~optimization. The~consolidated portfolio combines the~advantages of~the~two approaches used so far: it extends the~idea of~the~defaults, and~it is easy to~share such a~ranking of~subsequent algorithms. What is more, at the~same time, we take into account the~specifics of~a~given dataset using the~best configurations of~hyperparameters for similar data. The~proposed method does not require additional optimization, is parallelizable, and~has strong anytime model performance. This property is significant when we aim at good results with a~limited time budget. This makes \textit{consolidated learning} a~support for data scientists preparing entire collections of~models for similar prediction problems or other subsamples of~observations. In the~long run, applying \textit{consolidated learning} to~model deployment can significantly reduce optimization budgets.

\section{Related work}

Until now, it has been common for individuals to~use the~defaults implemented in the~software or to~use simple tuning methods. With the~development of~machine learning, more advanced hyperparameter optimization methods have been proposed; however, they require additional expertise in the~configuration itself. This is why some data scientists find them deterrent and~this is why they often neglect tuning. The~use of~random search methods was conducive to~automatic hyperparameter optimization gaining in popularity. Previously, it was known that the~configurations for many algorithms are crucial for the~performance of~trained models, but effective tuning of~the~settings was lacking. Random search facilitates determination of~a~low dimensional effective subspace of~hyperparameters faster than a~grid search or manual tuning, but it is still susceptible to~the~dimensionality of~the~searched space. To~eliminate low-efficiency configurations, faster multi-armed bandits methods such as~Successive Halving~\citep{jamieson2016sh} or Hyperband~\citep{li2017hyperband} are used. However, these more advanced methods work only for iterative algorithms and~are far less common than simple random search or defaults.

Bayesian-based optimization methods are a~class of~techniques that particularly require expert knowledge in implementation.
Their great advantage is using the~knowledge acquired from the~previous evaluations and~adjusting the~optimization process to~the~characteristics of~the~considered dataset. 
     However,  the~selection of~a~surrogate model is crucial for optimization effectiveness. The~most popular are variants of~SMBO \citep{jones1998efficient} such as~SMAC~\citep{hutter2011sequential}, TPE~\citep{bergstra2011algorithms} or Spearmint~\citep{snoek2012practical}. However, all of~them are computationally demanding, difficult to~parallelize and~depend on the~choice of~a~starting point~\citep{wistuba2015learning}. What is more, straightforward Bayesian-based methods do not provide anytime good solutions. To~eliminate this problem, adaptive resource allocation and~early-stopping of~unpromising configurations are combined with Bayesian optimization~\citep{falkner2018bohb}. Despite these modifications, we still do not leverage the~information that has been gathered so far in the~previous experiments for other datasets; the~only way to~provide additional information is the~prior distributions of~the~hyperparameters~\citep{oh2018bock,bayesianprior2021,perrone2019learning}.

In addition to~the~predefined portfolio of~hyperparameters employed in this article, there are different attempts to~combine the~strengths of~online and~offline approaches. The~most common is injection of~the~portfolio information into Bayesian optimization. The~main goal is to~exploit the~adaptability~of~online methods while leveraging offline portfolios in order to~quickly propose a~good, though perhaps not the~best, configuration of~hyperparameters. The~most common approach is to~define the~starting points in Bayesian optimization not as~random ones  but considering their model performance  in the~prior experiments~\citep{feurer2015initializing,feurer2019auto,wistuba2018scalable}.

    These methods emphasize adaptation of~the~surrogate model to~the~considered new dataset, and~the~portfolio is used only for initialization. An alternative is to~use the~results from the~previous experiments and~build a~black-box surrogate model that predicts the~performance for a~selected data set and~hyperparameter configuration. Then, based on the~data collected offline, we can make a~prediction of~the~response surface for the~new dataset~\citep{vilalta2004using,reif2014automatic,davis2018annotative,probst2019tunability}.

\section{Problem definition}

\subsection{Hyperparameter optimization}

Most machine learning algorithms are dependent on the~user specified hyperparameters. An algorithm is trained, i.e., values of~internal model parameters are updated iteratively, in accordance with the~chosen algorithm  and~the~data provided. So most machine learning algorithms $\mathcal{A}$  can be parametrized with dataset $D$ and~hyperparameter configuration $\lambda \in \Lambda \subset \mathbb{R}^d$. Dataset $D$ is a~finite sample from joint distribution $\DSp = (\XSp, \YSp)$, where $\XSp \subset \mathbb{R}^p$ is $p$-dimensional feature space and~$\YSp$ is target variable space, categorical or numerical. To~evaluate the~quality of~trained model  $\mathcal{A}(D,\lambda)$  we use quality function $F:\mathbb{D}\times \Lambda \to \mathbb{R}$ mapping a~dataset and~hypeparameters to~model performance, for instance, accuracy or an~area under curve (AUC). For every hyperparameter, we attempt to~estimate the~expected value of~performance  for a~random sample of~observations given as
\begin{equation}
\label{eq:generalizationF}
    G(\DSp, \lambda) = \mathbb{E}_{D \sim \DSp}\;F(D, \lambda).
\end{equation}
Since we observe only a~finite sample from $\DSp$ we estimate Equation~\ref{eq:generalizationF} using cross-validation or a~holdout subset of~data. 
The~main  objective of~hyperparameter optimization for prediction problem $\DSp$  is to~find  $ \lambda^*$ configuration, optimal in respect to~the~expected value of~model performance
\begin{equation}
\label{eq:optimization}
    \lambda^* = \argmax_{\lambda \in \Lambda}\; G(\DSp,\lambda).
\end{equation}

We consider different hyperparameter tuning strategies such as~the~mentioned earlier random search or Bayesian optimization to~find this configuration. However, these can not guarantee that we will find the~global maximum and~the~setting providing this, so we are interested in finding a~configuration that gives a~decent model performance, preferably after only a~few iterations.  

Most optimization strategies are based on a~trial of~the~finite sequence of~hyperparameter values  $\Lambda_T = (\lambda_1, \hdots, \lambda_T)$, where $T$ - the~number of~iterations depending on the~available budget. In random search, the~set $\Lambda_T $ is predefined and~independent of~the~prior experiments, and~every component $\lambda_i$ is sampled independently of~each other. In Bayesian-based methods $\Lambda_T $ is selected in runtime and~$\lambda_i$ is determined by a~surrogate model on the~base of~the~model performance for the~preceding values $\lambda_1, \hdots, \lambda_{i-1}$. 
 
 \subsection{Meta-learning in hyperparameter optimization}
 \label{sec:metalearningDef}
 One of~the~application of~meta-learning in hyperparameter optimization is transferring a~hyperparameter portfolio from a~set of~previously performed experiments to~a~new dataset. Like random search, a~predefined meta-learning portfolio is actually a~finite sequence of~configurations completely defined before tuning for the~considered new dataset. However, configurations $\lambda_i$ are not sampled independently but selected considering the~performance model from the~previously collected experiments, optimized for these experiments, and~then possibly extended to~randomly chosen new task. Herein lies the~primary assumption of~meta-learning and~hyperparameter transfer; we assume that configurations that worked for a~set of~previous tasks will also give a~decent performance for new, unknown data. The~algorithm by which the~portfolio is composed can vary, the~most commonly used is greedy search~\citep{wistuba2015learning}, but average ranking~\citep{brazdil2000comparison} can also be applied.  
 
In this article, the~main contribution is the~method of~determining the~family of~datasets for which the~optimization is performed, not the~procedure of~completing the~transferred portfolio of~hyperparameters. That is why we have to~highlight two sets of~tasks.  Firstly, we determine a~repository of~the~already tuned $N$ tasks and~call it meta-train set $\textbf{D}_{meta-train}$. Every dataset is associated with the~distribution $\DmetatrainSp{i}~=~ (\XmetatrainSp{i}, \YmetatrainSp{i})$ for $i=1,\hdots, N$, where $\XmetatrainSp{i} \subset \mathbb{R}^{p_i}$ is a~$p_i$ dimensional feature space.  Using $\textbf{D}_{meta-train}$ we will define the~meta-learned portfolio $\Lambda_T$. Secondly, a~meta-test task is not known before the~dataset for which we want solve the~Equation~\ref{eq:optimization}. The~meta-test task is sampled from distribution $\DmetatestSp~=~ (\XmetatestSp, \YmetatestSp)$ which is different than meta-train distribution. Let $P(\XmetatrainSp{i}), P(\XmetatestSp)$ be marginal probability distributions for  $i$-th meta-train prediction problem and~for meta-test respectively. Generally, every meta-train distribution $\XmetatrainSp{i}$ and~$\XmetatestSp$ is  defined independently of~each other, and~we do not assume any relationship between them. 
 
 As OpenML is a~major source of~prediction tasks, various unrelated datasets are used. For instance, one of~meta-train task \texttt{wdbc} describing prediction of~breast cancer consists of~numerical features extracted from image diagnostic and~the~ meta-test is \texttt{spambase} using word frequency statistics to~assess whether a~mail is spam. Every feature may come from markedly different sources and~distributions. However, even this kind of~meta-train selection ensures an~improvement in the~optimization strategy, especially in terms of~anytime performance. To~benefit from advantages of~such a~guided portfolio of~hyperparameters, we focus more on meta-train and~meta-test relationships in this work.

\subsection{Consolidated learning}

We define a~technique of~transferring the~hyperparameter from the~consolidated design meta-train set to~the~meta-test task as~\textit{consolidated learning}. The~motivation behind this modification comes from a~practical perspective on building predictive models.  

In real-world use cases, prediction tasks for specific domains often include standard explanatory variables shared between many datasets so that models can exploit analogous dataset characteristics. For example, according to~\citep{kumar2021machine}, $75\%$ of~analyzed articles concerning prediction of~Alzheimer disease dementia progression use cognitive assessments as~features. In extreme cases, data scientists working for a~specific entity often have one large database and~build multiple models for different data samples extracted, such as~\cite{koyner2018development} building a~sequence of~machine learning models to~predict an~acute kidney injury. One other case is the~need to~update the~model for samples from a~different time period. In that case, it is common to~update the~set of~observations and~train the~model anew. Such models use the~same set of~variables so the~models should have close properties to~the~previous versions. The~question remains how to~optimize the~new algorithm.

 These dependencies and~shared definitions of~features between meta-datasets is a~different scenario from what has been considered in research papers on meta-learning so far. To~capture these circumstances, from the~design of~the~sets used for training, we assume a~similarity between the~prediction problems. We expect that if machine learning algorithms can use similarly or identically distributed features then it should detect similar features interactions or treat  the~same common variable similarly. Since the~only parameterizations of~the~model that we know of~are hyperparameters we assume that such relationships between prediction tasks will positively affect the~transfer of~hyperparameters.

We formalize the~\textit{consolidated meta-train} and~\textit{consolidated learning} using the~terminology from Section~\ref{sec:metalearningDef}. Restricting meta-train tasks to~the~representative for specific domain results in dependency between $\XmetatrainSp{i}$ and~$\XmetatestSp$ for some $i=1, \hdots, N$. In particular, exploratory features with the~same marginal distribution may occur in two different meta-train and~meta-test distributions. In other words, some of~the~explanatory variables may be shared between the~two prediction problems under consideration. We term this situation as~$P(\XmetatrainSp{i}) \cap P(\XmetatrainSp{j}) \neq \emptyset$  for $i \neq j$ or  $P(\XmetatrainSp{i}) \cap P(\Xmetatest) \neq \emptyset$.  If features set is identical we denote this by $P(\XmetatrainSp{i}) \equiv P(\XmetatestSp) $. We define this constrained meta-train as~a~consolidated set in which common explanatory variables occur between the~sets contained in the~meta-train set and~the~meta-test set. On the~basis of~\textit{consolidated meta-training}, a~portfolio is composed (according to~any strategy) and~this process is called \textit{consolidated learning}.

Correspondence between \textit{consolidated meta-train} and~meta-test is significantly higher than between unrelated tasks within the~OpenML repository. The~assumption about shared variables allows us to~propose a~meta-feature-free strategy of~\textit{consolidated learning}, namely hyperparameter transfer which provides anytime solutions.

\section{metaMIMIC repository}
\label{section:database}

This section describes the~methodology for creating a~meta-dataset to~imitate the~\textit{consolidated learning} environment. Therefore, based on the~MIMIC-IV database \citep{johnson2020mimic} we create a~collection of~binary classification tasks of~varying similarity. We weigh three scenarios of~similarity between the~extracted tasks. In the~real world, such repositories are naturally collected during model development. However,to our knowledge, such a~repository is not available for research purposes. Behind the~choice of~the~MIMIC database as~a~source for the~collection of~prediction problems is its wide use in the~research for machine learning applications in medical diagnosis \citep{nemati2018interpretable,zhang2019machine,meng2021mimic,liu2021effects}. 
We employ this collection to~evaluate a~hyperparameter transfer in \textit{consolidated learning}  and~assess the~improvement in tunning.

\subsection{MIMIC-IV database}

MIMIC-IV (Medical Information Mart for Intensive Care) is an~extensive, freely available database comprising de-identified health-related data from patients admitted to~the~intensive care unit (ICU) of~the~Beth Israel Deaconess Medical Center. It contains data of~over 380,000 patients admitted to~the~ICU from 2008-2019. We include patient tracking data, demographics, laboratory measurements sourced from patient-derived specimens, and~information collected from the~clinical information system used during ICU stay. 

To determine the~cohort selection, we have to~define the~patient inclusion criteria taking into account machine learning principles \citep{DBLP:conf/mlhc/JohnsonPM17,meng2021mimic,DBLP:journals/jbi/PurushothamMCL18}.  We consider only the~first admission of~every patient to~preserve the~independence of~all observations. Every patient must be at least 15 years old at the~time of~hospitalization and~his admission must correspond to~at least one chart event, one lab event, and~one diagnosis recorded in the~database.
The~hospital stay length must be shorter than 60 days. In total, 34925 unique patients met all the~above conditions.

\subsection{Prediction tasks}

To determine multiple predictive problems, we decided to~predict the~occurrence of~a~specific disease. We examined 50 most commonly appearing conditions and~hand-picked groups of~diseases that have a~representation in both ICD-9 and~ICD-10 codes (see Table \ref{tab:selected_targets}). It resulted in 12 targets for binary classification. We also considered whether the~selectedtargets  can be successfully predicted with the~data available in the~MIMIC-IV database (at least 0.7 mean ROC AUC in 4-fold cross-validation after tuning).

\begin{table}[!ht]
	\caption{Selected targets with corresponding ICD codes and~frequency in the~considered cohort.}
	\centering
	\begin{tabular}{llll}
		\toprule
		Condition                                       & ICD-9        & ICD-10  & Frequency \\
		\midrule
		Hypertensive diseases                           & 401-405      & I10-I16 & 59.8\%    \\
		Disorders of~lipoid metabolism                  & 272          & E78     & 40.3\%    \\
		Anemia                                          & 280-285      & D60-D64 & 35.9\%    \\
		Ischematic heart disease                        & 410-414      & I20-I25 & 32.8\%    \\
		Diabetes                                        & 249-250      & E08-E13 & 25.3\%    \\
		Chronic lower respiratory diseases              & 466, 490-496 & J40-J47 & 19.5\%    \\ 
		Heart failure                                   & 428          & I50     & 19.4\%    \\
		Hypotension                                     & 458          & I95     & 14.5\%    \\
		Purpura and~other hemorrhagic conditions        & 287          & D69     & 11.9\%    \\ 
		Atrial fibrillation and~flutter                 & 427.3        & I48     & 10.5\%    \\
		Overweight, obesity and~other hyperalimentation & 278          & E65-E68 & 10.5\%    \\
		Alcohol dependence                              & 303          & F10     & 7.7\%     \\
		\bottomrule
	\end{tabular}
	\label{tab:selected_targets}
\end{table}

We selected 58 features, hand-picking ones with the~lowest number of~missing values.
Besides gender and~age, we included only numerical variables related to~the~purposeful medical examination. Most features were recorded several times, so we aggregated them a~minimum, average, and~maximum values. In total, this resulted in 172 variables. The~missing values are imputed with a~mean of~all observations for each task independently to~avoid data leakage.

\subsection{Task correspondence}
\label{sec:scenarios}

In addition to~specifying the~response variables and~the~explanatory variable space, we also considered various assumptions about choosing the~subset of~observations and~available variables. Generally, in applications such choices are forced by the~available data, such as~the~size of~the~sample of~observations that can be used, or how model validation is defined. 

\begin{figure}[!ht]
    \centering
    \includegraphics[scale = 0.55]{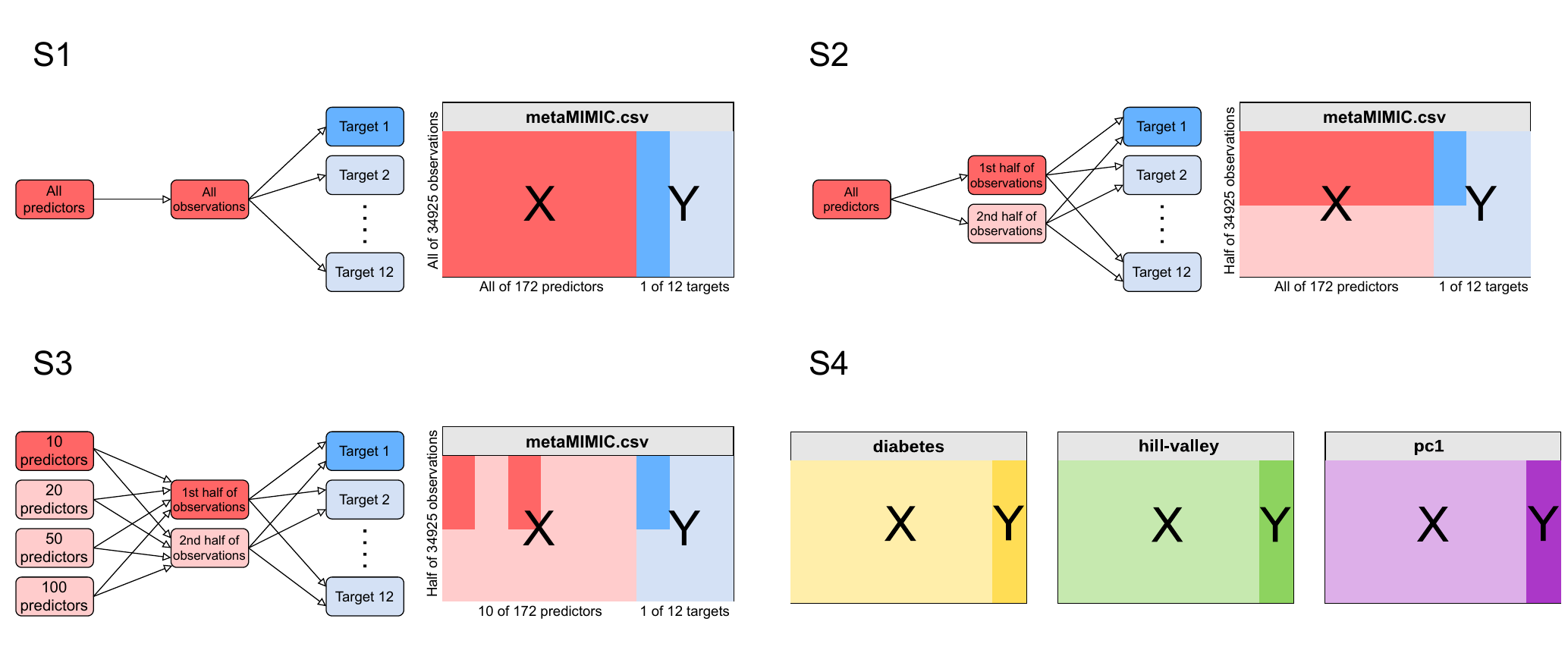}
    \caption{Schemas of~design decision in creating scenarios of~similarity between meta-train and~meta-test. In \Sone-\Sthree  from MIMIC-IV we extracted feature space, sample of~observations and~target disease. In \Sone models for meta-train and~meta-test use all predictors and~observations but targets vary. Scenario \Stwo contain models built for the~same predictors but disjoint sample of~observations. In \Sthree we consider different subset of~predictors.  In \Sfour meta-train is composed of~the~OpenML datasets unrelated to~MIMIC.}
    \label{fig:diagramScenarios}
\end{figure}

We mimic different selection scenarios that affect the~intuitive perception of~similarity between the~obtained tasks in this work. The~process of~task design always consists of~three choices -- which predictors to~use, which observations to~consider and~which target to~predict (see Figure~\ref{fig:diagramScenarios}). To~verify the~impact of~similarity between the~tasks on the~\textit{consolidated learning}, we compare them with baseline transfer from a~wide range of~OpenML datasets, in addition to~the~three scenarios of~task correspondence.

\begin{enumerate}
    \item In the~first scenario (\Sone), we predict different targets considering the~same observations and~using the~same variables. Using formal notation feature space are identical $\XmetatrainSp{i}~\equiv~\XmetatestSp$ for every~$i = 1, \hdots, N$. Therefore, the~only real choice to~make is to~select which target to~predict (Figure~\ref{fig:diagramScenarios} \Sone). This setup reflects a~situation where these targets are determined by historical data of~the~hospital's patients comprising basic diagnostic tests and~diagnoses. The~only difference between the~tasks is the~diagnosis we want to~predict, so there are $12$ prediction sets.
    
    \item In the~second scenario (\Stwo), various targets are predicted considering different samples observations but using the~same set of~variables. To~avoid leakage of~information occurring in \Sone, we consider two random, disjoint samples of~observations (Figure \ref{fig:diagramScenarios} \Stwo) but the~models are provided with the~same $172$ variables. This experimental setting corresponds to~the~situation where we consider models built on out-of-time samples of~patients but from the~same distribution. When considering any two prediction problems, we can examine models built on independent sets of~observations. In this setup, we get $2\times 12 = 24$ prediction tasks.
    
    \item In the~third scenario (\Sthree),  we manipulate not only observations samples but also a~set of~variables to~predict defined tasks. We select a~different number of~the~most important predictors for each task (Figure~\ref{fig:diagramScenarios} \Sthree). The~choice of~predictor set  was realized through selecting top $n$ variables with the~highest permutation variable importance value \citep{breiman2001random,DBLP:journals/jmlr/FisherRD19}, calculated using the~XGBoost model with default settings. We determine $n =10, 20, 50, 100$ out of~$172$ features. This scenario imitates the~transfer of~knowledge between models built upon different targets, but  now the~scenario takes into account not identical 
    feature space.Many predictive problems are based on an~core set of~variables and~these are available in many tasks. An example is blood tests performed and~used in the~diagnosis of~most diseases. So when considering a~broad class of~medical problems, many of~the~tasks contain variables describing such measurements. But there are also more specific tests for example cognitive testing is the~primary diagnosis of~Alzheimer's or ECG for heart diseases. If we are considering sets of~models that predict these diseases then the~datasets will have corresponding variables.
 In this setup we get $4 \times 2\times 12=96$ prediction tasks. 
    
    \item As a~baseline meta-set and~meta-learning approach (\Sfour), we use 22 datasets from the~OpenML repository. Using this collection for meta-learning, even in an~online approach, has proven better than using random search or uninformed Bayesian optimization.
    \end{enumerate}

\section{Experiment methodology}
\label{section:methodology}

The~proposed method of~model tuning for the~MIMIC-IV database is based on hyperparameters transfer  within  a~collection of~medical prediction problems. We validate the~effectiveness of~using a~MIMIC family of~prediction problems by comparing analogous tuning strategies determined for an~unrelated family of~datasets with the~OpenML.

\subsection{Hyperparameter grid}
\label{sec:grid}

As a~hyperparameter search space, we use the~grid from the~MementoML study~\citep{Kretowicz2020MementoML} to~validate the~\textit{consolidated learning} with the~results obtained from 22 machine learning tasks from the~OpenML repository. The~designed grid comprises 1000 sets of~8 different XGBoost hyperparameters sampled independently from the~ predefined distributions. The~considered hyperparameters and~the~distributions they  are sampled from are presented in Table \ref{tab:parameters}. If \texttt{gblinear}  is selected  as~a~booster, not all hyperparameters are active. 

The~predefined grids of~hyperparameters exemplifies discretization of~the~searched  space. However, the~predefined grid approach has been used in several works on optimization~\citep{wistuba2015learning,Wistuba2016SequentialTuning} so we decided to~create a~fixed random grid. It uses the~advantages of~random search and~allows efficient space search while ensuring the~reproducibility of~results. 

\begin{table}[!ht]
	\caption{Hyperparameters and~their underlying distributions. $U$ stands for a~random variable sampled from a~uniform distribution with corresponding lower and~upper bounds. Booster can be either \texttt{gblinear} or \texttt{gbtree} with equal probability. With \mbox{*} we indicate the~active hyperparameters  when booster = \texttt{gblinear}. }
	\centering
	\begin{tabular}{lllllc}
		\toprule
		& Hyperparameter     & Type     & Lower & Upper & Distribution  \\
		\midrule
		\mbox{*}& n\_estimators      & integer  & 1     & 1000  & U            \\
		\mbox{*}   & learning\_rate     & float    & 0.031 & 1     & $2^U$    \\
		\mbox{*}  & booster            & discrete & -     & -     &   \{\texttt{gblinear}, \texttt{gbtree}\}          \\
		& subsample          & float    & 0.5   & 1     & U        \\
		& max\_depth         & integer  & 6     & 15    & U       \\
		& min\_child\_weight & float    & 1     & 8     & $2^U$       \\
		& colsample\_bytree  & float    & 0.2   & 1     & U      \\
		& colsample\_bylevel & float    & 0.2   & 1     & U        \\
		\bottomrule
	\end{tabular}
	\label{tab:parameters}
\end{table}  

For each task in scenarios \Sone - \Sthree, we train XGBoost models for a~given grid of~hyperparameters using 4-fold cross-validation (CV). In scenario \Sfour we use results from MementoML. ROC AUC is used as~model performance measure.
Due to~the~incomparability of~AUC values between the~tasks,  mean 4-CV ROC AUC is scaled to~interval $[0,1]$ for each task individually.

\subsection{Tuning strategies}
We test four hyperparameter tuning methods for the~XBGoost algorithm. Two of~them are meta-learning based, so they use the~performance model results for other meta-train sets. For each scenario, the~following strategies are tested: the~transfer is performed with metaMIMIC  (\Sone-\Sthree) and~OpenML (\Sfour) independently. In the~hyperparameter transfer within the~metaMIMIC, we used one-task-out validation. For scenario \Sfour, we studied the~transfer of~hyperparameters configuration into the~optimization for metaMIMIC tasks from scenario \Stwo. As a~meta-learning strategy of~formation portfolio we considered:

\begin{itemize}
    \item Average Sequential Model Free Optimization (A-SMFO) \citep{Wistuba2016SequentialTuning} using the~greedy algorithm to~determine a~sequence of~hyperparameters to~test on the~new task. The~order in the~hyperparameter portfolio is initially optimized for the~meta-train set and~the~best configurations for each meta-train dataset are included. In the~consecutive iterations, we add configurations among the~feasible candidates, not considering the~ previously chosen ones. This offline algorithm aims to~create a~diverse configuration portfolio covering a~wide range of~prediction problems. 

\item  Average Ranks Ranking Method (AR) \citep{brazdil2000comparison} determining the~order of~hyperparameters according to~the~average ranking obtained by configurations for every meta-train dataset. This method is elementary and~does not require additional computations.

\end{itemize}

 Both meta-learning methods are limited to~hyperparameter configuration derived from the~grid defined in Section~\ref{sec:grid}.
 As a~baseline tuning strategy, we tested random search and~Bayesian optimization as~two strategies not exploiting results from other datasets. 
 \begin{itemize}
     \item Random search (RS) is simulated as~a~random walk within a~defined hyperparameters grid. Due to~this, we did not perform a~random search several times to~estimate the~expected learning curve and~its variance; we could figure the~theoretically expected model performance after $t$  iterations determined by the~expected value of~the~beta distribution with the~relevant parameters and~empirical parameters quantiles.
     \item Bayesian optimization (BO) is performed using the~implementation available in the~\texttt{scikit-optimize} package based on uniform distributions of~hyperparameters, with bounds corresponding to~the~MementoML grid. Since Bayesian optimization may propose different configurations of~hyperparameters than on our given grid of~hyperparameters, it also validates the~quality of~the~proposed search space. 
 \end{itemize}

Since A-SMFO, AR, and~RS use evaluation from the~same hyperparameter grid for every task, the~observed maximum of~AUC measure is the~same for all the~ three tuning strategies. They only vary in a~sequence of~proposed configurations. Bayesian optimization is not limited to~this grid only and~can find better or worse optimal AUC than the~other optimizations.

\subsection{Evaluation of~tunning strategies}

The~objective of~the~experiment is to~see if we can improve hyperparameter tuning for one dataset from metaMIMIC using meta-learning for scenarios~\Sone~-~\Sthree relative to~\Sfour. Let us recall that we use a~one-task-out schema for scenarios \Sone-\Sthree to~build the~meta-train set. Furthermore, to~avoid information leakage for scenarios \Stwo and~\Sthree, we always exclude the~meta-learning-based optimization strategy, the~target for which we optimize. For example, let us consider scenario \Stwo and~diabetes target variable  with the~first  subsample of~observations in meta-train. We include only the~tasks for the~second  subsample of~observations but exclude the~tasks for diabetes.
For scenario \Sfour, the~OpenML dataset repository is independent of~the~metaMIMIC, so we do not address this problem.

We compare the~optimization strategies for all scenarios and~each dataset individually. For every iteration of~optimization of~a~given strategy, we consider the~best performance obtained so far. That is why, we are interested in reaching the~maximal value as~soon as~possible and~in learning which strategy achieves this. To~aggregate this information for the~entire collection of~datasets, we recorded the~development of~the~average rank among different hyperparameter tuning strategies. Furthermore, to~assess the~speed of~convergence to~observed optimum, we use the~average distance (ADTM) to~maximum AUC value~\citep{wistuba2015learning}.

Let $\textbf{D}_{meta-test}$ be the~collection of~meta-test datasets for which tunning strategy is evaluated, and~$f$ is AUC measure. The~portfolio of~hyperparameter configuration proposed by strategy in iteration $T$ is $\Lambda_T$. Then ADTM is defined~as
\begin{equation}
    ADTM(\textbf{D}_{meta-test}, \Lambda_T) = \frac{1}{	\lvert  \textbf{D}_{meta-test}\rvert} \sum_{i \in \textbf{D}_{meta-test}} \min_{ \lambda \in \Lambda_T} \frac{f_i^{max} - f_i(\lambda) }{f_i^{max}- f_i^{min}},
\end{equation}
where $ \lvert \textbf{D}_{meta-test}\rvert $ is a~cardinality of~meta-test set. In our experiment setup as~meta-test we consider all MIMIC-based tasks available in examined scenario \Sone-\Sfour. For each meta-test task portfolio is determined by corresponding meta-train, according to~one-dataset-out validation.

\begin{figure}[!ht]
    \centering 
    \includegraphics[width=0.75\textwidth]{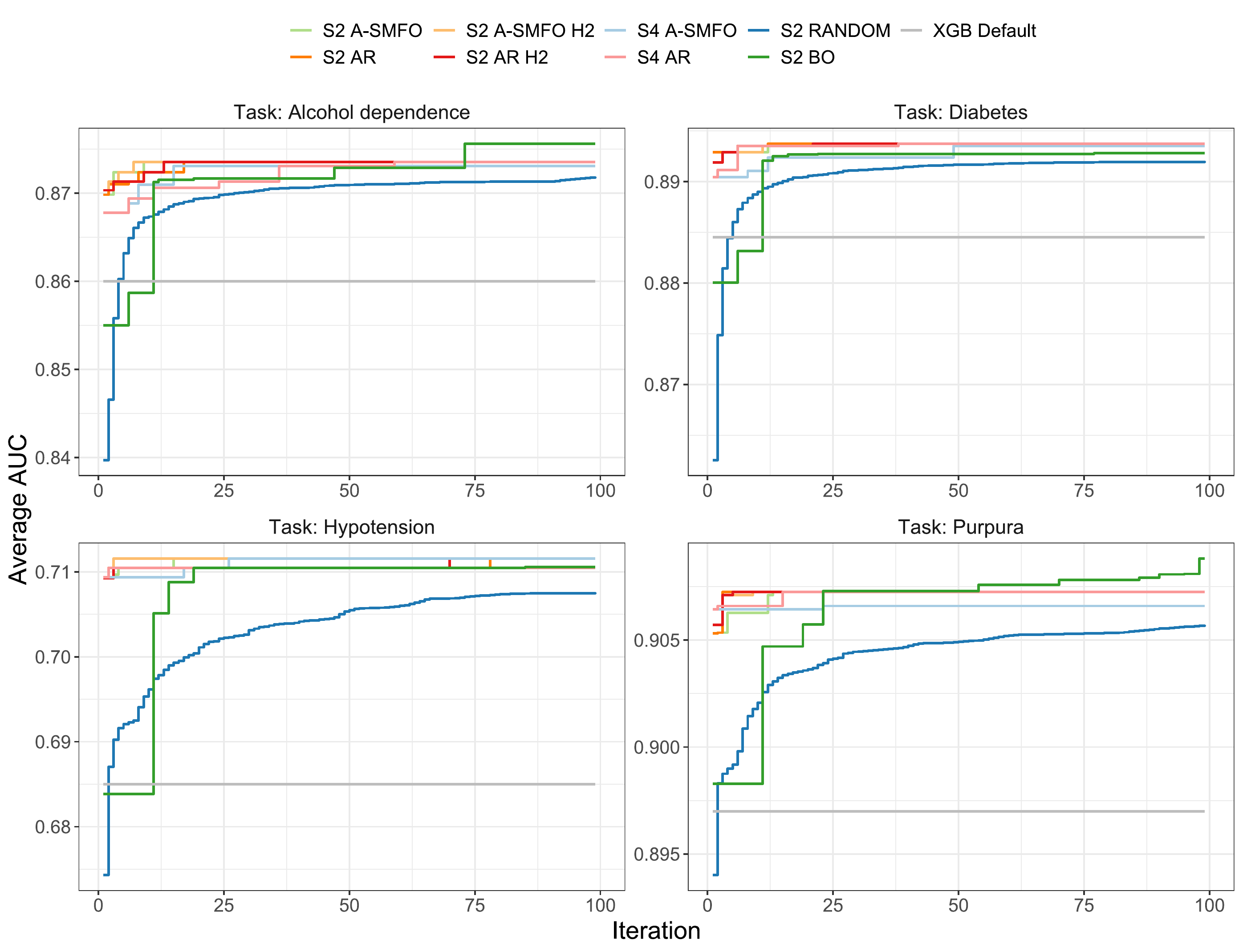}
    \caption{Hyperparameter tuning velocity of~different methods and~multiple tasks. Purpura is the~only task for which OpenML initially significantly outperforms MIMIC-IV among the~12 tasks considered.}
    \label{fig:lines}
\end{figure}


\section{Effectiveness of~consolidated learning }
\label{sec:results:tuning}

To assess the~improvement in transferability of~hyperparameters brought by \textit{consolidated learning} we perform optimization for every metaMIMIC task from Scenario \Stwo. We build portfolios upon the~meta-train in Scenarios \Stwo and~\Sfour. For \Stwo, we consider the~meta-train including the~same  sample of~observations and~a~disjoint sample of~observations as~in the~meta-test task. We also test two strategies for creating a~static portfolio, A-SMFO, and~AR. The~baseline collates meta-learning results with random search, Bayesian optimization, and~default XGBoost algorithm.

 Figure \ref{fig:lines} shows the~results for the~four selected meta-test targets. Looking at the~model performance during hyperparameter tuning, we see that methods based on meta-learning  provide configurations close to~the~observed maximum already in the~first iteration. The~distance between the~learning curves for scenarios \Stwo, \Sfour, and~the~baselines is evident for the~first few iterations. These results are significantly better than the~model without tuning. Despite being able to~go beyond the~specified grid, Bayesian optimization obtains better results for only two targets, so we may assume that the~predefined grid covers the~space of~good configurations.

Let us take a~closer look at the~meta-learning-based methods. The~differences in model performance between the~transfer for \textit{consolidated learning} in scenario \Stwo and~the~transfer based on OpenML are of~the~order of~$10^{-3}$, but in most cases, domain-based strategies reach maximum AUC before OpenML-based methods. Furthermore, the~difference between the~transfer in scenario \Stwo regarding the~identical and~disjoint sample of~observations is negligible, so the~effect of~the~subset of~observations is not significant. A-SMFO and~AR strategies  of~building a~portfolio give close learning curves even in the~first iteration.

\begin{figure}[!hb]
    \centering
    \includegraphics[scale = 0.4]{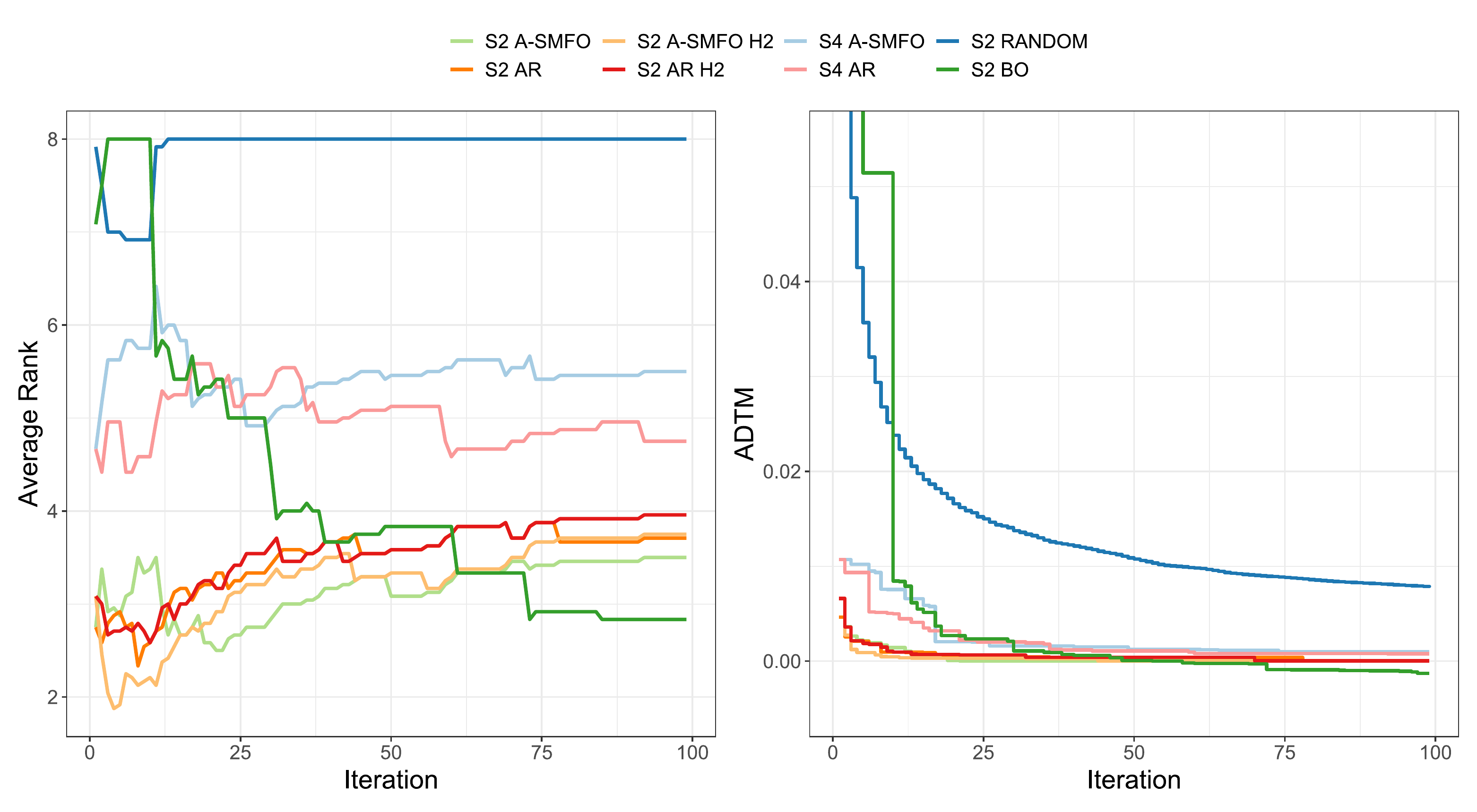}
    \caption{Comparison of~the~aggregated performance development with the~increasing number of~iterations for different optimization strategies. In the~left plot, changes in the~average rank of~strategy are used to~summarize overall efficiency. In the~right, ADTM is applied. As meta-test tasks always are used MIMIC-based task from scenario \Stwo. For meta-train, we use tasks from \Stwo for the~disjoint subset of~observations (S2 A-SMFO, S2 AR), the~same subset of~observations (S2 A-SMFO H2, S2 AR H2). As baseline strategies, we use meta-train from scenario \Sfour (S4 A-SMFO, S4 AR), random search (S2 RANDOM), and~Bayesian optimization (S2 BO). }
    \label{fig:rankADTM}
\end{figure}

To further summarize the~impact of~strategy selection between different tasks and~scenarios, we examine the~change in average rankings for each strategy (Figure~\ref{fig:rankADTM}). We see that in the~earlier iterations, portfolios from \Stwo \textit{consolidated learning} get a~better rank than the~configurations extracted from OpenML, especially in the~early iterations. All strategies based on \textit{consolidated learning} have a~similar average rank, regardless of~a~subsample of~observations and~a~method of~creating a~portfolio algorithm. Only Bayesian optimization exceeds the~consolidated optimization but requires about 30 iterations to~approach the~\Stwo strategies. In Figure~\ref{fig:rankADTM} we see how fast the~tuning strategies  converge against the~best hyperparameter configuration on average.
Similarly, we observe the~learning curves for \textit{consolidated learning} converge considerably faster than the~other strategies. Again, this marked difference is more substantial in the~OpenML meta-data set. As the~rank of~each strategy changes with time, we see that all line associated with \Stwo scenario converge  to~the~observed maximum the~fastest. OpenML-based strategies are slightly worse AUC but reach the~maximum after about 10 iterations. Bayesian optimization goes beyond the~fixed-parameter grid, and~to~see if it finds better hyperparameters than those included in the~grid, ADTM for this optimization is computed assuming that the~maximum observed value is equal to~the~maximum observed on the~predefined grid. Hence, negative ADTM values for BO over 75 interactions.

Thus, we can conclude that meta-learning is effective and~even using unrelated datasets allows us to~reject unsatisfactory configurations and~provide a~decent model performance for several trials in tunning. We can accelerate the~tunning by employing \textit{consolidated learning.} Random search and~Bayesian optimization need to~go through several iterations to~achieve comparable results as~methods based on a~static portfolio created from the~previous experiments.

\section{Robustness of~transferability}

In Section~\ref{sec:results:tuning}, we saw that meta-learning-based methods find the~optimum observed on the~defined grid after only 10 iterations. In this section, we explore the~similarity between hyperparameter spaces in model performance terms. We also investigate the~effect of~task correspondence on the~strength of~hyperparameter transfer. This is especially important in order to~provide anytime solution for optimization.

\begin{figure}[!ht]
    \centering 
    \includegraphics[width=\textwidth]{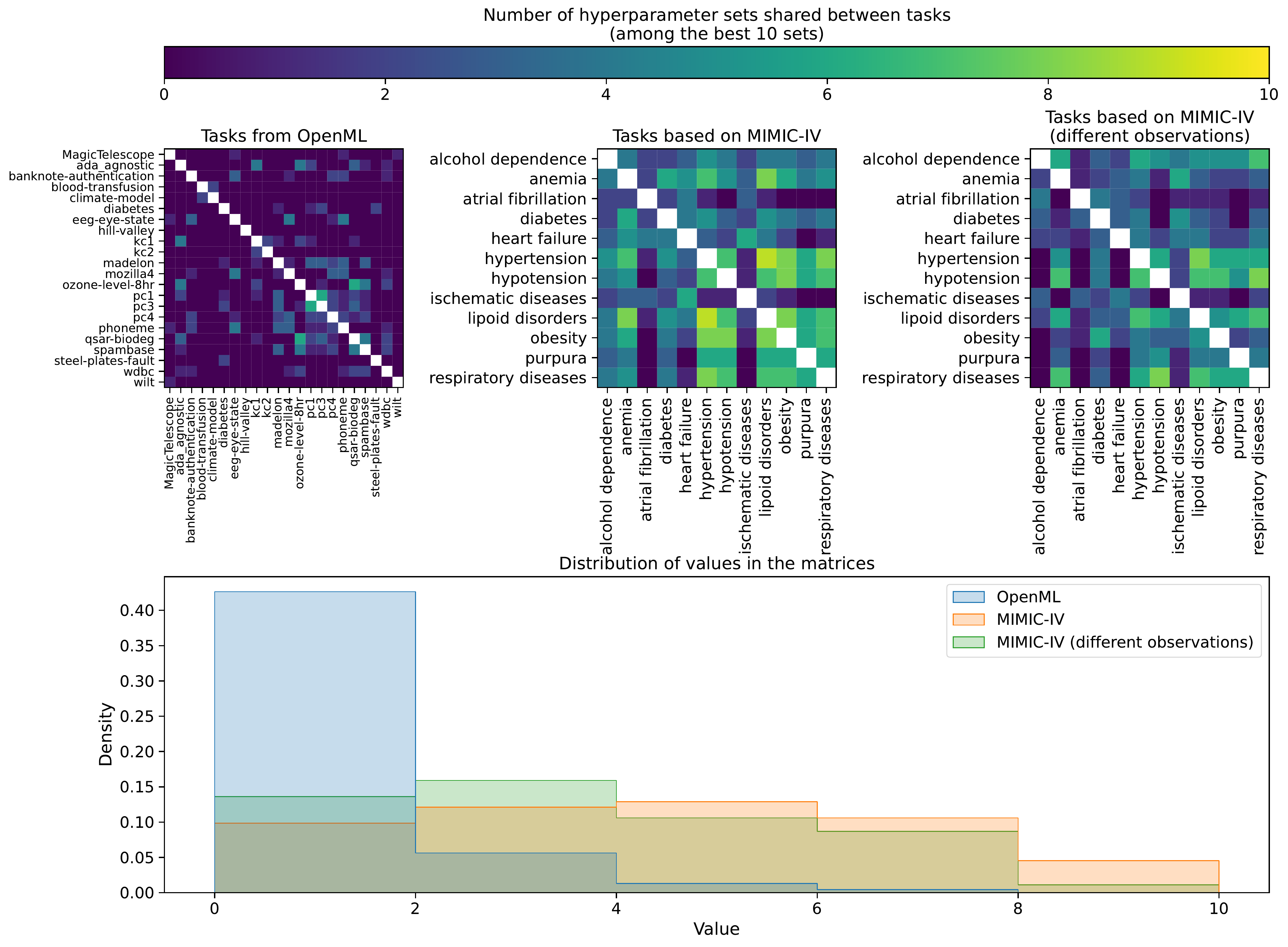}
    \caption{Numbers of~the~best 10 hyperparameter sets (regarding the~mean 4-CV ROC AUC measure) shared between tasks from \Sfour, \Sone, and~\Stwo. An individual cell of~the~matrix corresponds to~the~number of~hyperparameter sets shared between a~given pair of~tasks. Histograms summarize the~distribution of~the~values of~each matrix. White color on the~diagonal means that the~value is not considered.}
    \label{fig:matrix_1}
\end{figure}

To analyze the~consistency of~the~hyperparameters model performance between any pair of~tasks, we examine the~percentage of~overlap in the~top 10 best configurations (Figure \ref{fig:matrix_1}). We decide on a~threshold of~1\% because $10$ iterations in tuning is sufficient for strategies \Stwo and~\Sfour. Nevertheless, choosing another threshold value from a~reasonable range of~10-100 results in analogous relationships between the~distributions of~values in the~matrices. This fact is also reflected in the~mean of~Spearman rank correlation coefficients calculated for individual pairs of~full rankings (0.165 ± 0.469 for \Sfour, 0.885 ± 0.072 for \Sone, and~0.849 ± 0.078 for \Stwo).

Comparing the~distributions of~values in the~presented matrices shows that the~number of~shared best hyperparameter sets is significantly higher for the~\Sone  than for the~\Sfour scenario  representing a~meta-learning from unrelated problems. In addition, with the~rightmost matrix, it is apparent that considering disjoint subsets of~observations (which is often closer to~actual use cases) results in only a~slight decrease in the~average number of~shared hyperparameter sets. 

\begin{figure}[!ht]
    \centering 
    \includegraphics[width=0.7\textwidth]{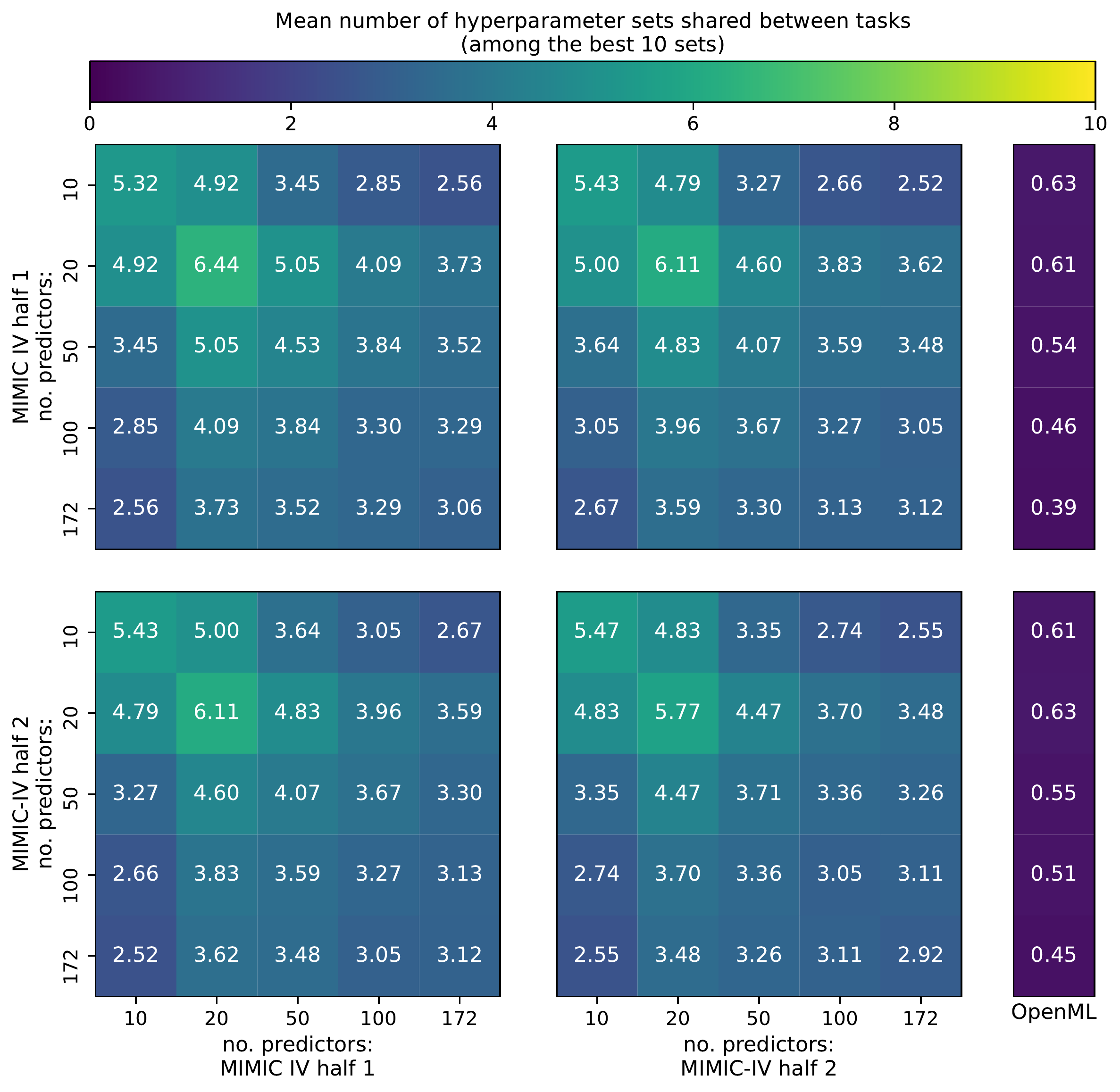}
    \caption{Summary of~mean numbers of~the~best ten hyperparameter sets (regarding the~mean 4-CV ROC AUC measure) shared between tasks from \Sthree. A single matrix cell represents the~average value for tasks with a~given number of~columns and~is based on a~given subset of~observations. Additionally, the~vectors on the~right correspond to~the~same operation for the~intersection of~MIMIC-IV-based tasks with tasks derived from OpenML.}
    \label{fig:matrix_2}
\end{figure}

The~analysis of~the~results scenario \Sthree required a~partial aggregation of~the~calculated statistics because without this operation, the~number of~possible combinations would become too large for clear representation in a~graph. We decided to~perform this aggregation by grouping the~tasks based on their source and, for MIMIC-IV, also on the~number of~predictors and~the~considered subset of~observations (Figure \ref{fig:matrix_2}). Therefore, a~single cell corresponds to~the~average value of~a~matrix created in the~same way as~the~previous graph.

As the~number of~predictors decreases, their diversity between the~tasks increases, which is due to~the~procedure of~selecting them described in Section~\ref{sec:scenarios}. Despite this, the~average number of~shared hyperparameter sets for tasks based on a~similar number of~predictors is consistently high. This suggests that \textit{consolidated learning} is related to~the~transferability of~hyperparameters. Nonetheless, even in the~worst case, the~average number of~shared hyperparameter sets is higher between the~pairs of~MIMIC-IV-based tasks than when intersecting the~MIMIC-IV-based tasks with the~tasks derived from the~OpenML.


\section{Conclusions}

The~results presented in this work highlight the~importance of~selecting meta-training repositories. To~our knowledge, this is the~first work analyzing the~impact of~meta-train sets on the~optimization power of~predefined hyperparameter portfolios. We show that purposefully accumulating results from the~prior prediction problems described by similar sets of~variables strengthens the~optimization strategies. We demonstrate empirically that leveraging datasets from the~MIMIC database produces better model performance than using a~portfolio determined for a~diverse repository. We observe a~positive effect of~application of~\textit{consolidated learning} both in tunning speed and~the~consistency of~the~best hyperparameters. We also analyze the~weakening of~assumptions in simulated \textit{consolidated learning} - despite  smaller constraints in individual \textit{consolidated learning} scenarios, we still show a~more significant transfer than for OpenML datasets. 

Using the~database MIMIC-IV, we demonstrate how \textit{consolidated meta-train} repositories can be constructed in practice. To~our knowledge, this is the~first approach to~creating a~domain-specific repository for meta-learning. It is worth noting that the~defined data dependency problem captures the~real use of~that database in both academic work and~practical model deployment. What is more, the~conducted research uses non-synthetic data from a~real-world source but allows us to~simulate the~different relationships between meta-train and~meta-test.

Our approach enhances the~meta-learning effect in hyperparameter optimization while avoiding the~problem of~defining a~representative set of~meta-features. This approach attempts to~answer whether hyperparameter transfer occurs at all and~whether it can be correlated with some definitions of~meta-features. If the~transfer does not occur for such restrictively defined tasks, it is hard to~imagine that we are able to~define meta-features that catch the~similarity of~predictive problems. Based on this study, we see that the~transfer is more evident within MIMIC-IV-based tasks.

In future work, we plan to~verify the~hypothesis that the~consolidated portfolios created for the~experiments extracted from MIMIC-IV give better performance for disease prediction problems based on the~history collected during hospital admission. This may be the~first step, leading to~domain-specific portfolios for a~broader class of~problems than defined in this work and~transferring hyperparameters between different problems without requiring  the~datasets to~partially share the~same variable definitions.

The~code needed to~reproduce the~metaMIMIC and~the~whole study can be found in this repository: \href{https://github.com/ModelOriented/metaMIMIC}{https://github.com/ModelOriented/metaMIMIC}.

\section*{Acknowledgments}
The~work on this paper is financially supported by the~NCN Opus grant 2017/27/B/ST6/01307.

\bibliographystyle{apalike}
\bibliography{references}  

\end{document}